\documentclass{article}

\usepackage{spconf}

\usepackage{graphicx}
\graphicspath{{fig/}}

\usepackage{algorithm}  
\usepackage{algpseudocode}  

\usepackage{amsmath}
\usepackage{amssymb}

\DeclareMathOperator*{\argmin}{arg\,min}

\renewcommand{\vec}[1]{\boldsymbol{\mathbf{#1}}}
\newcommand{\defeq}{\triangleq}

\usepackage{booktabs}
\usepackage{tabularx}
\usepackage{multirow}
\usepackage{siunitx}

\newcolumntype{C}{>{\centering\arraybackslash}X}
\newcolumntype{L}{>{\raggedright\arraybackslash}X}
\newcolumntype{R}{>{\raggedleft\arraybackslash}X}
\newcommand{\beforecaption}{\vspace*{-5pt}}
\newcommand{\aftercaption}{\vspace*{7.5pt}}

\usepackage{microtype}
\usepackage{url}

\usepackage{balance}
\usepackage{cite}
\title{An embedded segmental K-means model for \\ unsupervised segmentation and clustering of speech}
\name{\ldots}
\name{Herman Kamper$^1$, Karen Livescu$^2$, Sharon Goldwater$^3$}
\address{
    $^1$Electrical and Electronic Engineering, Stellenbosch University, South Africa\\
    $^2$Toyota Technological Institute at Chicago, United States\\
    $^3$ILCC, School of Informatics, University of Edinburgh, United Kingdom\\
    {\small \tt kamperh@sun.ac.za, klivescu@ttic.edu, sgwater@inf.ed.ac.uk}
    }

\usepackage{todonotes}
\usepackage{xcolor}
\definecolor{orange}{HTML}{FF6600}

\begin{document}


\maketitle

\begin{abstract}
Unsupervised segmentation and clustering of unlabelled speech are core problems in zero-resource speech processing. Most approaches lie at methodological extremes: some use probabilistic Bayesian models with convergence guarantees, while others opt for more efficient heuristic techniques. Despite competitive performance in previous work, the full Bayesian approach is difficult to scale to large speech corpora. We introduce an approximation to a recent Bayesian model that still has a clear objective function but improves efficiency by using hard clustering and segmentation rather than full Bayesian inference. Like its Bayesian counterpart, this embedded segmental \textit{K}-means model (ES-KMeans) represents arbitrary-length word segments as fixed-dimensional acoustic word embeddings.   We first compare ES-KMeans to previous approaches on common English and Xitsonga data sets (5 and 2.5 hours of speech): ES-KMeans outperforms a leading heuristic method in word segmentation, giving similar scores to the Bayesian model while being 5 times faster with fewer hyperparameters. However, its clusters are less pure than those of the other models. We then show that ES-KMeans scales to larger corpora by applying it to the 5 languages of the Zero Resource Speech Challenge 2017 (up to 45 hours), where it performs competitively compared to the challenge baseline.\footnote{\url{https://github.com/kamperh/recipe_zs2017_track2}}
\end{abstract}


\section{Introduction}

The growing area of \textit{zero-resource speech processing} aims to develop unsupervised methods that can learn directly from raw speech audio in settings where transcriptions, lexicons and language modelling texts are not available.
Such methods are crucial for providing speech technology in 
languages where transcribed data are hard or impossible to collect, e.g., unwritten or endangered languages~\cite{besacier+etal_speechcom14}.
In addition, such methods may shed light on how human infants acquire language~\cite{rasanen_speechcom12,dupoux_arxiv16}.

Several zero-resource tasks have been studied, including acoustic unit discovery~\cite{varadarajan+etal_acl08,lee+glass_acl12,ondel+etal_icassp17}, unsupervised representation learning~\cite{badino+etal_icassp14,renshaw+etal_interspeech15,zeghidour+etal_interspeech16}, query-by-example search~\cite{zhang+etal_icassp12,levin+etal_icassp15} and topic modelling~\cite{gish+etal_interspeech09,kesiraju+etal_icassp17}.
Early work mainly focused on unsupervised term discovery, where the aim is to automatically
find repeated word- or phrase-like patterns in a collection of speech~\cite{park+glass_taslp08,jansen+vandurme_asru11,oosterveld+etal_icassp17}.
While useful, the discovered patterns are typically isolated segments spread out over the data, leaving much speech as background.
This has prompted several studies on \textit{full-coverage} approaches, where the entire speech input is segmented and clustered into word-like units~\cite{sun+vanhamme_csl13,walter+etal_asru13,lee+etal_tacl15,kamper+etal_taslp16,elsner+shain_emnlp17}.

Two such full-coverage systems have recently been applied to the data of the Zero Resource Speech Challenge 2015 (ZRSC'15), giving a useful basis for comparison~\cite{versteegh+etal_sltu16}.
The first is the Bayesian embedded segmental Gaussian mixture model (BES-GMM)~\cite{kamper+etal_arxiv16}: a probabilistic model that represents potential word segments as fixed-dimensional acoustic word embeddings, and then builds a whole-word acoustic model in this embedding space while jointly doing segmentation.
The second is the recurring syllable-unit segmenter (SylSeg)~\cite{rasanen+etal_interspeech15}, a cognitively motivated, fast, heuristic method that applies unsupervised syllable segmentation and clustering and then predicts recurring syllable sequences as words.
These two models are representative of two methodological extremes often seen in zero-resource systems: either probabilistic Bayesian models with convergence guarantees  are used~\cite{vanhainen+salvi_interspeech12,vanhainen+salvi_icassp14,ondel+etal_icassp17,lee+etal_tacl15,kamper+etal_arxiv16,taniguchi+etal_tcds16}, or heuristic techniques are used in pipeline approaches~\cite{walter+etal_asru13,rasanen+etal_interspeech15}.

Here we introduce an approximation to BES-GMM that falls in between these two extremes.
The \emph{embedded segmental \textit{K}-means model} (ES-KMeans) uses hard clustering and segmentation, rather than full Bayesian inference.
Nevertheless, it has a clear objective function, in contrast to heuristic methods such as SylSeg.
Compared to BES-GMM, it has fewer hyperparameters and a simpler optimization algorithm since probabilistic sampling is not necessary; ES-KMeans is therefore more efficient, while still having a principled objective.

Hard approximations have been used since the start of probabilistic modelling in supervised speech recognition~\cite{wilpon+rabiner_assp85,rabiner+etal_att86,juang+rabiner_assp90}, and also in more recent work to improve the efficiency of an unsupervised Bayesian model~\cite{shum+etal_taslp16}.
We are therefore following in a long tradition of using hard approximation. However, all of these studies applied it in frame-by-frame modelling approaches, while our approach operates on embedded representations of whole speech segments.
There is a growing focus on such \textit{acoustic word embedding} methods~\cite{levin+etal_asru13,levin+etal_icassp15,kamper+etal_icassp16,chung+etal_interspeech16,settle+livescu_slt16,he+etal_iclr17,audhkhasi+etal_icassp17,settle+etal_interspeech17}, since they make it possible to easily and efficiently compare variable-duration speech segments in a fixed-dimensional space.

We conduct two sets of experiments.
First, we analyze how {the hard approximations of ES-KMeans} 
affect speed and accuracy relative to the original BES-GMM and the SylSeg method.
On English and Xitsonga data {(used in ZRSC'15)}, we show that ES-KMeans outperforms SylSeg in word segmentation and gives similar scores to BES-GMM, while being 5 times faster.
However, the cluster purity of ES-KMeans falls behind that of the other two models.
We show that the higher purity for BES-GMM results from a tendency towards smaller clusters which, unlike in ES-KMeans, can also be varied using hyperparameters.
In the second set of experiments---conducted as part of the Zero Resource Speech Challenge 2017 (ZRSC'17)---we show that ES-KMeans can also scale to larger corpora by applying it to the data sets of up to 45 hours for 5 languages.

\section{Embedded segmental K-means}

Starting from standard \textit{K}-means, we describe the embedded segmental \textit{K}-means (ES-KMeans) objective~and~algorithm.

\subsection{From \textit{K}-means to ES-KMeans objective~function}

Given a speech utterance consisting of acoustic frames $\vec{y}_{1:M} = \vec{y}_1, \vec{y}_2, \ldots, \vec{y}_M$ (e.g., MFCCs), our aim is to break the sequence up into word-like segments, and to cluster these into hypothesized word types.

\begin{figure}[tb]
    \centering
  \includegraphics[width=0.9\linewidth]{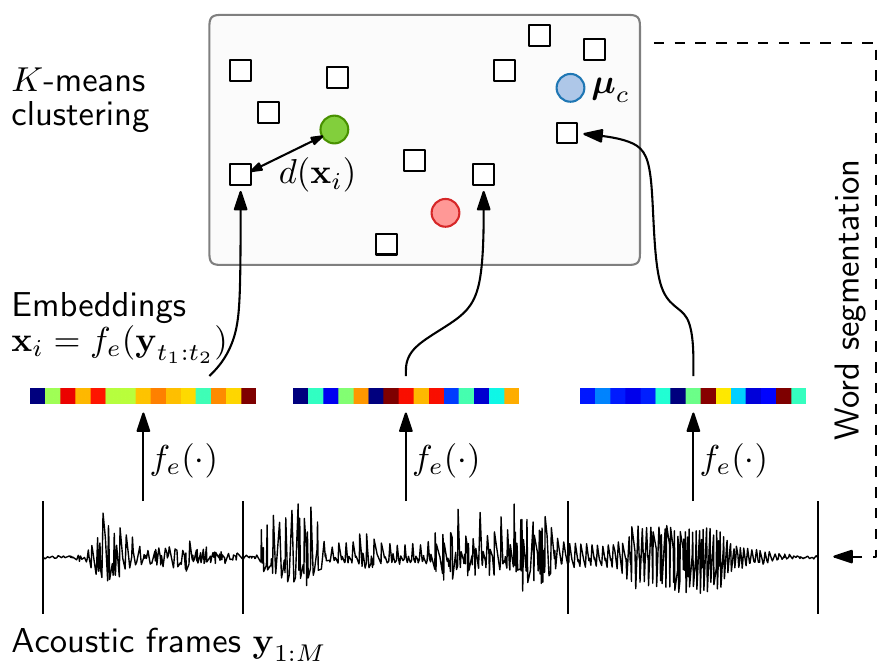}
    \beforecaption
    \caption{The embedded segmental \textit{K}-means model for unsupervised segmentation and clustering of speech.}
    \label{fig:segmental_kmeans}
\end{figure}

If we knew the segmentation (i.e., where word boundaries occur),
the data would consist of several segments of different durations, as shown at the bottom of Fig.~\ref{fig:segmental_kmeans}.
To cluster these, we need a method to compare variable-length vector sequences.
One option would be to use an alignment-based distance measure, such as dynamic time warping.
Here we instead follow an~\textit{acoustic word embedding} approach~\cite{levin+etal_asru13,levin+etal_icassp15,kamper+etal_icassp16}: 
an embedding function $f_e$ is used to map a variable length speech segment to a single embedding vector $\vec{x} \in \mathbb{R}^D$ in a fixed-dimensional space, i.e., segment $\vec{y}_{t_1:t_2}$ is mapped to a vector $\vec{x}_i = f_e (\vec{y}_{t_1:t_2})$, illustrated as coloured horizontal vectors.
The idea is that speech segments that are acoustically similar should lie close together in $\mathbb{R}^D$, allowing segments to be efficiently compared directly in the embedding space without alignment.
{Various such embedding methods have been proposed, ranging from graph-based~\cite{levin+etal_asru13} to unsupervised recurrent neural approaches~\cite{chung+etal_interspeech16}.
Here we use a very simple method: any segment is uniformly downsampled so that it is represented by the same fixed number of vectors, which are then flattened to obtain the embedding~\cite{levin+etal_asru13}.
But note that ES-KMeans is agnostic to the embedding method, so future improvements in embeddings can be incorporated~directly.}

Embedding all segments in the data set would give a set of vectors $\mathcal{X} = \{ \vec{x}_i \}_{i = 1}^N$, which could be clustered into $K$ hypothesized word classes using \textit{K}-means, as shown at the top of Fig.~\ref{fig:segmental_kmeans}.
Standard \textit{K}-means aims to minimize the sum of squared Euclidean distances to each cluster mean:
$\min_{\vec{z}} \sum_{c = 1}^K \sum_{\vec{x} \in \mathcal{X}_c} || \vec{x} - \vec{\mu}_c ||^2$, where $\left\{ \vec{\mu}_c \right\}_{c = 1}^K$ are the cluster means, $\mathcal{X}_c$ are all vectors assigned to cluster $c$, and element $z_i$ in $\vec{z}$ indicates which cluster $\vec{x}_i$ belongs to.
The standard algorithm alternates between reassigning vectors to the closest cluster means, and then updating the means.

Standard \textit{K}-means would be appropriate if the segmentation was known, but this is not the case in this zero-resource setting.
Rather, the embeddings $\mathcal{X}$ can change depending on the current segmentation.
For a data set 
of $S$ utterances, 
we denote the segmentations as $\mathcal{Q} = \{ \vec{q}_i \}_{i = 1}^S$, 
where $\vec{q}_i$ indicates the boundaries for utterance $i$.
$\mathcal{X}(\mathcal{Q})$ is used to denote the embeddings under the current segmentation.
Our aim now is to jointly optimize the cluster assignments $\vec{z}$ and the segmentation $\mathcal{Q}$.  Under what objective should these be optimized?

One option would be to extend the standard \textit{K}-means objective and optimize $\min_{(\mathcal{Q} ,\, \vec{z})} \sum_{c = 1}^K \, \sum_{\vec{x} \in  \mathcal{X}_c \cap  \mathcal{X} (\mathcal{Q})} || \vec{x} - \vec{\mu}_c ||^2$, where $\mathcal{X}_c \cap \mathcal{X}(\mathcal{Q})$ are embeddings assigned to cluster $c$ under segmentation $\mathcal{Q}$.
But this is problematic: imagine inserting no boundaries over an utterance, resulting in a single embedding and a single term in the summation; any other segmentation would result in more terms in the summation, likely giving an overall worse score---even if all embeddings
are close to cluster means.
Therefore, instead of assigning a score \textit{per segment}, we assign a score \textit{per frame}.
This score is given by the score achieved by the segment to which that frame belongs, implying that segment scores are weighed by duration:
\begin{equation}
    \min_{\mathcal{Q} \,,\, \vec{z}} \sum_{c = 1}^K \, \sum_{\vec{x} \in {\mathcal{X}_c} \cap \mathcal{X}(\mathcal{Q})} \textrm{len}(\vec{x}) \, || \vec{x} - \vec{\mu}_c ||^2
    \label{eq:segmental_kmeans}
\end{equation}
where $\textrm{len}(\vec{x})$ is the number of frames in the sequence on which embedding $\vec{x}$ is calculated.

The overall ES-KMeans algorithm initializes word boundaries randomly, and then optimizes~\eqref{eq:segmental_kmeans} by alternating between optimizing segmentation $\mathcal{Q}$ while keeping cluster assignments $\vec{z}$ and means $\left\{ \vec{\mu}_c \right\}_{c = 1}^K$ fixed (top to bottom in Fig.~\ref{fig:segmental_kmeans}), and then optimizing the cluster assignments and means while keeping the segmentation fixed (bottom to top in the figure).

\subsection{Segmentation}

Under a fixed clustering $\vec{z}$, the objective~\eqref{eq:segmental_kmeans} becomes
\begin{align}
    \min_{\mathcal{Q}} \sum_{\vec{x} \in \mathcal{X}(\mathcal{Q})} \textrm{len}(\vec{x}) \, || \vec{x} - \vec{\mu}^*_{\vec{x}} ||^2 = \min_{\mathcal{Q}} \sum_{\vec{x} \in \mathcal{X}(\mathcal{Q})} d(\vec{x})
    \label{eq:kmeans_segmentation_objective}
\end{align}
where $\vec{\mu}^*_{\vec{x}}$ is the mean of the cluster to which $\vec{x}$ is currently assigned (according to $\vec{z}$), and 
$d(\vec{x}) \defeq \textrm{len}(\vec{x}) \, || \vec{x} - \vec{\mu}^*_{\vec{x}} ||^2$ is the ``score'' of embedding $\vec{x}$ (lower $d$ is better).

Eq.~\eqref{eq:kmeans_segmentation_objective} can be optimized separately for each utterance: we want to find the segmentation $\vec{q}$ for each utterance  that gives the minimum of the sum of the scores of the embeddings under that segmentation.
This is exactly the problem addressed by the shortest-path algorithm (Viterbi) which uses dynamic programming to solve this problem efficiently~\cite[$\S$21.7]{erickson_algorithms}.

Let $q_t$ 
be
the number of
frames in the hypothesized segment (word) that ends at
 frame $t$: if $q_t = j$, then $\vec{y}_{t - j + 1:t}$ is a word.\footnote{For a an utterance with frames $\vec{y}_{1:M}$, a sequence of $q$'s ending with $q_M$ specifies a unique segmentation; boldface $\vec{q}$ is this sequence of $q$'s.}
We define forward variables 
$\gamma[t]$ as the optimal score up to boundary position $t$: $\gamma[t] \defeq \min_{q_{:t}} \sum_{\vec{x} \in \mathcal{X}(q_{:t})} d(\vec{x})$, with $q_{:t}$ the sequence of segmentation decisions (the $q$'s) up to $t$.
These can be recursively calculated~\cite[$\S$21.7]{erickson_algorithms}:
\begin{align}
    \gamma[t]
    &= \min_{j = 1}^t \left\{ d\left( f_e(\vec{y}_{t - j + 1:t}) \right) + \gamma[t - j] \right\}
    \label{eq:gamma_1}
\end{align}
Starting with $\gamma[0] = 0$, we calculate~\eqref{eq:gamma_1} for $1 \leq t \leq M - 1$.
We keep track of the optimal choice ($\argmin$) for each $\gamma[t]$, and the overall optimal segmentation is then given by starting from the final position $t = M$ and moving backwards, repeatedly choosing the optimal~boundary.

\subsection{Cluster assignments and mean updates}

\noindent For a fixed segmentation $\mathcal{Q}$, the objective~\eqref{eq:segmental_kmeans} becomes
\begin{equation}
    \min_{\vec{z}} \sum_{c = 1}^K \sum_{\vec{x} \in \mathcal{X}_c \cap \mathcal{X}(\mathcal{Q})} \textrm{len}(\vec{x}) || \vec{x} - \vec{\mu}_c ||^2
\end{equation}
When the means $\left\{ \vec{\mu}_c \right\}_{c = 1}^K$ are fixed, the optimal reassignments \eqref{eq:kmeans_assign} follow standard \textit{K}-means, and are guaranteed to improve \eqref{eq:segmental_kmeans} since the distance between the embedding and its assigned cluster mean never increases:
\begin{equation}
    z_i = \argmin_c \left\{ \textrm{len}(\vec{x}_i) \, || \vec{x}_i - \vec{\mu}_c ||^2 \right\} = \argmin_c || \vec{x}_i - \vec{\mu}_c ||^2
    \label{eq:kmeans_assign}
\end{equation}

\noindent Finally, we fix the assignments $\vec{z}$ and update the means:
\begin{equation}
    \vec{\mu}_c
    = \frac{1}{\sum_{\vec{x} \in \mathcal{X}_c} \textrm{len}(\vec{x})} \sum_{\vec{x} \in \mathcal{X}_c} \textrm{len}(\vec{x}) \, \vec{x} 
    \approx \frac{1}{N_c} \sum_{\vec{x} \in \mathcal{X}_c} \vec{x} \label{eq:approx_mean}
\end{equation}
The exact equation is the mean of the vectors assigned to cluster $c$ weighed by duration, and is guaranteed to improve~\eqref{eq:segmental_kmeans}.
We use the approximation, which is exact if all segments have the same duration, to again match standard \textit{K}-means, with $N_c$ the number of embeddings currently assigned to cluster $c$.

The complete ES-KMeans algorithm is given below.
Since the segmentation, clustering and mean updates each improve~\eqref{eq:segmental_kmeans}, the algorithm will converge to a local optimum.

\begin{algorithm}[!h]
\caption{The embedded segmental \textit{K}-means algorithm.}\label{alg:segmental_kmeans}
\begin{algorithmic}[1]
\small
\State Initialize segmentation $\mathcal{Q}$ randomly.
\State Initialize cluster assignments $\vec{z}$ randomly.
\Repeat \Comment{Optimization iterations}

    \For{$i = $ randperm$(1$ to $S)$} \Comment{Select utterance $i$}
       
        \State Calculate $\gamma$'s using~\eqref{eq:gamma_1}. \Comment{Segmentation variables}

        \State $t \gets M_i$ 
        \While{$t \geq 1$} \Comment{Perform segmentation}
            \State $q_t \gets \argmin_{j = 1}^t \left\{ d\left( f_e(\vec{y}_{t - j + 1:t}) \right) + \gamma[t - j] \right\}$
            \State $t \gets t - q_t$
        \EndWhile
        
        \State Assign new embeddings $\mathcal{X}(\vec{q}_i)$ to clusters using~\eqref{eq:kmeans_assign}.

        \State Update means using~\eqref{eq:approx_mean}.

    \EndFor
\Until{convergence}
\end{algorithmic}
\end{algorithm}
\vspace*{-2.5pt}

\subsection{The Bayesian embedded segmental GMM}
\label{sec:segmental_bayesian_gmm}

In previous work~\cite{kamper+etal_taslp16,kamper+etal_arxiv16}, we proposed a very similar model, but instead of \textit{K}-means, we used a Bayesian GMM as whole-word clustering component (top of Fig.~\ref{fig:segmental_kmeans}).
This Bayesian embedded segmental GMM (BES-GMM) served as inspiration for ES-KMeans;
we briefly discuss their relationship here.

A Bayesian GMM treats its mixture weights $\vec{\pi}$ and component means $\left\{ \vec{\mu}_c \right\}_{c = 1}^K$ as random variables rather than point estimates, as is done in a regular GMM.
We use conjugate priors: a Dirichlet prior over $\vec{\pi}$ and a spherical-covariance Gaussian prior over $\vec{\mu}_c$.
All components share the same fixed covariance matrix $\sigma^2 \vec{I}$.
The model is then formally defined as:

\vspace*{-6pt}
\noindent
\begin{minipage}{.45\linewidth}
    \centering
    \begin{alignat}{2}
        &\vec{\pi}  &&\sim \textrm{Dir}\left( a/{K} \vec{1} \right) \label{eq:fbgmm1} \\
        &z_i &&\sim \vec{\pi}  \label{eq:fbgmm2}
    \end{alignat}
    ~
\end{minipage}
\hfill
\begin{minipage}{.45\linewidth}
    \centering
    \begin{alignat}{2}
        &\vec{\mu}_c  &&\sim \mathcal{N} (\vec{\mu}_0, \sigma_0^2 \vec{I})  \label{eq:fbgmm3} \\
        &\vec{x}_i &&\sim \mathcal{N} (\vec{\mu}_{z_i}, \sigma^2 \vec{I})  \label{eq:fbgmm4}
    \end{alignat}
    ~
\end{minipage}

\vspace{-10pt}\noindent Under this model, component assignments and a segmentation can be inferred jointly using a collapsed Gibbs sampler~\cite{resnik+hardisty_gibbs_tutorial10}.
Full details are given in~\cite{kamper+etal_taslp16}, but the Gibbs sampler looks very similar to Algorithm~\ref{alg:segmental_kmeans}: the Bayesian GMM gives likelihood terms (``scores'') in order to find an optimal segmentation, while the segmentation hypothesizes boundaries for the word segments which are then clustered using the GMM.
However, for BES-GMM, component assignments and segmentation are sampled probabilistically, instead of making hard decisions.

The link between the two models emerges asymptomatically.
It can be shown that standard \textit{K}-means results from a GMM as the variances approach zero~\cite[$\S$20.3.5]{barber},~\cite{kulis+jordan_icml12}.
In a similar way it can be shown that the Gibbs sampling equations for segmentation and component assignments for BES-GMM (as given in~\cite{kamper+etal_taslp16}) approach~\eqref{eq:gamma_1} and~\eqref{eq:kmeans_assign}, respectively, in the limit $\sigma^2 \rightarrow 0$, when all other hyperparameters are fixed.

Without giving a full complexity analysis, we note that because ES-KMeans only considers the closest cluster, it is more efficient than BES-GMM, where all components are considered when assigning embeddings to clusters and during segmentation (since embedding ``scores'' are obtained by marginalizing over all components).
ES-KMeans can also be trivially parallelized, since both segmentation and cluster assignment can be performed in parallel for each utterance. This parallelized algorithm is still guaranteed to converge, 
 though possibly to a different local optimum than Algorithm~\ref{alg:segmental_kmeans} since updates are in a different order.
Parallelizing BES-GMM is also possible, but the guarantee of converging to the true posterior distribution is lost~\cite{shum+etal_taslp16}.

\subsection{Heuristic recurring syllable-unit word segmentation}

We also compare to
the ZRSC'15 submission of R{\"a}s{\"a}nen et al.~\cite{rasanen+etal_interspeech15}.
Their
system, which we refer to as SylSeg, relies on a novel cognitively motivated unsupervised method that predicts boundaries for syllable-like units, and then clusters these units on a per-speaker basis.
Using a bottom-up greedy mapping, recurring syllable cluster sequences are~then~predicted~as~words.

SylSeg is much simpler in terms of computational complexity and implementation than ES-KMeans or BES-GMM.
But, in contrast to the heuristic methodology followed in SylSeg, both ES-KMeans and BES-GMM have clear overall objective functions that they optimize, the one using hard clustering, the other using Bayesian inference.

\begin{table*}[tbp]
    \caption{(a)~Performance of models on the two test corpora. Lower NED is better.  Runtimes for SylSeg$^*$ are rough estimates, obtained from personal communication with the authors~\cite{rasanen+etal_interspeech15}. (b)~English development set performance (\%) of BES-GMM as the variance is~varied.}
    \aftercaption
    \label{tbl:results}
    (a)~
    \begingroup
        \small
        \renewcommand{\arraystretch}{1}
        \begin{tabularx}{0.65\linewidth}{@{}l@{}CcCCCCC@{}}
            \toprule
            & Cluster purity & WER & NED & Boundary \textit{F} & Token\ \ \ \ \textit{F} & Type\ \ \ \ \ \ \ \textit{F} & Runtime (s) \\
            \midrule
            \textit{English (\%)} \\
            SylSeg & - & - & 71.1 & 55.2 & 12.4 & 12.2 & 100$^*$ \\
            ES-KMeans & 42.8 & 73.2 & 71.6 & \textbf{62.2} & \textbf{18.1} & \textbf{18.9} & {193} \\
            BES-GMM & \textbf{56.1} & \textbf{68.3} & \textbf{55.5} & \textbf{62.2} & 17.9 & 18.6 & 1052 \\ 
            \midrule
            \textit{Xitsonga (\%)} \\
            SylSeg & - & - & 62.8 & 33.4 &2.7  &  3.3 & 20$^*$ \\ 
            ES-KMeans & 40.5 & 80.3 & 70.4 & 42.1 & 3.7 & 4.9 & \textbf{44} \\ 
            BES-GMM & \textbf{49.8} & \textbf{71.6} & \textbf{58.4} & \textbf{43.1} & \textbf{4.0} & \textbf{5.2} & 196 \\ 
            \bottomrule
        \end{tabularx}
%
    \endgroup
    ~\hfill(b)
    \begingroup
        \small
        \renewcommand{\arraystretch}{1.1}
        \begin{tabularx}{0.25\linewidth}{lCc}
            \toprule
            \multicolumn{1}{c}{$\sigma^2$} & Cluster purity & WER \\
            \midrule
            0.00001 & 56.1 & 68.9 \\
            0.0001 & 55.7 & 69.0 \\
            0.001 & \textbf{56.9} & \textbf{67.7} \\
            0.0015 & 51.9 & 69.8\\
            0.00175 & 41.1 & 75.8 \\
            0.002 & 35.1 & 86.8 \\
            \bottomrule
        \end{tabularx}
    \endgroup
\end{table*}

\section{Experiments}
\label{sec:experiments}

{We perform two sets of experiments.
First, we compare ES-KMeans to SylSeg and BES-GMM on the ZRSC'15 data.
SylSeg and BES-GMM have both previously been applied to the relatively small ZRSC'15 corpora, making it ideal for a comparative analysis.
Second, we apply ES-KMeans to the ZRSC'17 data.
Although there are no previous results to compare to, the ZRSC'17 data sets are considerably larger than those of ZRSC'15,  providing a useful test of scalability.
}

\subsection{Experimental setup and evaluation}
\label{sec:experimental_setup}

As in~\cite{kamper+etal_taslp16,ludusan+etal_lrec14,kamper+etal_arxiv16}, we use several metrics to evaluate against ground truth forced alignments.
By mapping every discovered word token to the ground truth token with which it overlaps most and then mapping every cluster to its most common word, average cluster purity and unsupervised word error rate (WER) can be calculated.\footnote{We allow more than one cluster to be mapped to the same word.}
By instead mapping every token to the true phoneme sequence with which it overlaps most, the normalized edit distance (NED) between all segments in the same cluster can be calculated; lower NED is better, with scores from 0 to 1.
Word boundary {precision, recall and} \textit{F}-score evaluates segmentation performance by comparing proposed and true word boundaries; similarly, word token {precision, recall and} \textit{F}-score measures the accuracy of proposed word token intervals.
Word type {precision, recall and} \textit{F}-score compares the set of unique phoneme mappings (obtained as for NED) to the set in the true lexicon.
See~\cite{ludusan+etal_lrec14} for full details.
{Cluster purity and WER are not considered in the ZRSC, so for some systems these results are not reported below.}

Our implementation of ES-KMeans follows as closely as possible that of BES-GMM in~\cite{kamper+etal_arxiv16}.
Both use uniform downsampling as embedding function $f_e$: a segment is represented by flattening 10 equally spaced MFCCs, with suitable interpolation. 
Both models use unsupervised syllable pre-segmentation~\cite{rasanen+etal_interspeech15} to limit allowed word boundaries.
For BES-GMM we use simulated annealing, an all-zero vector for $\vec{\mu}_0$, $\sigma_0^2 = \sigma^2/\kappa_0$, $\kappa_0 = 0.05$, $a = 1$ and $\sigma^2 = 0.001$. 

\subsection{Results: Comparison to other systems and analysis}
\label{sec:results_zs2015}

{In the first set of experiments, we use the two ZRSC'15 data sets: an English corpus of around 5 hours of speech from 12 speakers, and a Xitsonga corpus of 2.5 hours from 24 speakers~\cite{versteegh+etal_interspeech15}.
We also use a separate English set of 6 hours for development.
In order to compare to previously published results~\cite{rasanen+etal_interspeech15,kamper+etal_arxiv16}, all systems here are applied in a speaker-dependent setting, and results are averaged over speakers.
As in~\cite{kamper+etal_arxiv16}, for both ES-KMeans and BES-GMM, $K$ is set to 20\% of the number of first-pass segmented syllables. Word candidates are limited to span at most 6 syllables, and must be at least 200~ms in duration.
We do not parallelize ES-KMeans here, in order to keep it as close as possible to BES-GMM.
}

Table~\ref{tbl:results}(a) shows the performance of the three models on the English and Xitsonga corpora.
Some of the SylSeg scores are unknown since these were not part of the {ZRSC'15} evaluation~\cite{rasanen+etal_interspeech15}.
Compared to BES-GMM, ES-KMeans achieves worse purity, WER and NED, but similar boundary, token and type \textit{F}-scores.
This comes with a 5-time improvement in runtime.
ES-KMeans achieves worse NED than SylSeg, but much better word boundary, token and type \textit{F}-scores.
SylSeg, however, is twice as fast.

\begin{figure}[tb]
    \centering
    \centerline{\includegraphics[width=0.85\linewidth]{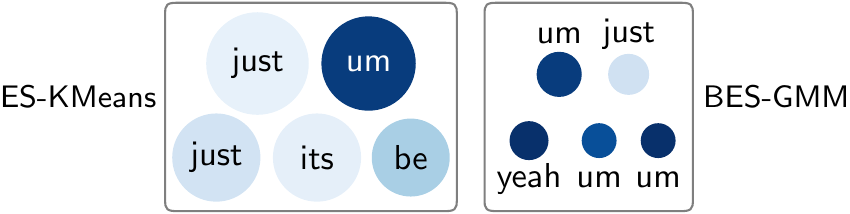}}
    \beforecaption
    \caption{The 5 biggest clusters for ES-KMeans and BES-GMM. Circle radii are according to cluster size; shading indicates purity. The cluster-to-true-word mapping is also shown.}
    \label{fig:clusters}
\end{figure}

ES-KMeans is therefore competitive in terms of word segmentation scores (boundary, token \textit{F}-scores) and lexicon quality (type \textit{F}-score), but falls behind in the purity-based metrics (purity, WER, NED).
The difference with BES-GMM is particularly interesting since $\sigma^2$ is set to be quite small, and ES-KMeans results from BES-GMM in the limit $\sigma^2 \rightarrow 0$.
To understand the discrepancy in purity, we analyzed ES-KMeans and BES-GMM on a single English development speaker.

\begin{figure}[tb]
    \vspace*{-3pt}
    \centering
    \centerline{\includegraphics[width=\linewidth]{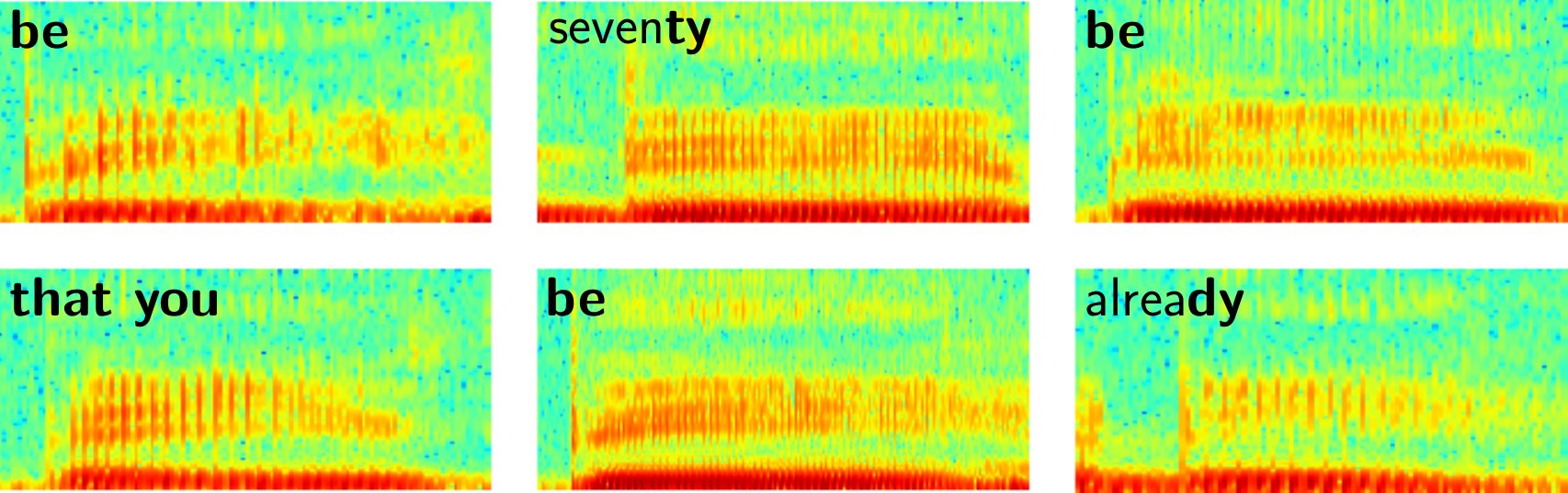}}
    \beforecaption
    \caption{Spectrograms for random tokens from the cluster of ES-KMeans mapped to ``be'' in Fig.~\ref{fig:clusters}. The portion of each true word which is covered by the segment is shown in bold.}
    \label{fig:spectrograms}
\end{figure}

For a qualitative view, Fig.~\ref{fig:clusters} shows the 5 biggest clusters for the two models.
BES-GMM outputs more smaller clusters with a higher purity (often separating the same word over different clusters) compared to ES-KMeans.
By
listening to tokens assigned to the same cluster by ES-KMeans, we found that, although tokens overlap with different
ground truth labels,
cluster assignments are qualitatively sensible, capturing  similarity in acoustics or prosody.
E.g., Fig.~\ref{fig:spectrograms} shows spectrograms for tokens assigned to the ``be'' cluster in Fig.~\ref{fig:clusters}. The ground truth word labels with maximal overlap are also shown.
For the ``seventy'' and ``already'' tokens, the segments only cover part of the true words (bold), and the ``that you'' token is actually pronounced in context as \mbox{[dh uw]}.
So despite mapping to different true labels, these segments form a reasonable acoustic group.
Nevertheless, they are penalized under purity~and~WER.

By spreading out its discovered tokens more evenly over clusters (Fig.~\ref{fig:clusters}), BES-GMM produces a clustering that is better-matched to the evaluation metrics, although the ES-KMeans clustering might be subjectively reasonable.
This spreading (or sparsity) of BES-GMM can be controlled through the fixed spherical covariance parameter $\sigma^2$, which impacts both the soft assignments of an embedding to a cluster and the segmentation ($\S$\ref{sec:segmental_bayesian_gmm}).
Table~\ref{tbl:results}(b) shows performance on the development set as $\sigma^2$ is varied. There is a sweet spot: when $\sigma^2$ is too big, most tokens are sucked up by a number of large garbage clusters; when $\sigma^2$ is smaller, more tokens are assigned to separate clusters.
In contrast, ES-KMeans has no $\sigma^2$ parameter and considers \textit{only} the single closest cluster.

\begin{table*}[tbp]
    \caption{Performance of the baseline system and ES-KMeans on the five languages of Zero Resource Speech Challenge 2017.}
    \centering
    \aftercaption
    \begingroup
    \small
    \label{tbl:results_zs2017}
    \renewcommand{\arraystretch}{1}
    \begin{tabularx}{0.85\linewidth}{@{}lRrrrrrrrrrrc@{}}
        \toprule
        &              &     & \multicolumn{3}{c}{Boundary} & \multicolumn{3}{c}{Token} & \multicolumn{3}{c}{Type} & Runtime \\
        \cmidrule{4-6} \cmidrule(l){7-9} \cmidrule(l){10-12} \cmidrule(l){13-13}
        & Coverage & \multicolumn{1}{c}{NED} & \multicolumn{1}{c}{Prec.} & \multicolumn{1}{c}{Rec.} & \multicolumn{1}{c}{\textit{F}} & \multicolumn{1}{c}{Prec.} & \multicolumn{1}{c}{Rec.} & \multicolumn{1}{c}{\textit{F}} & \multicolumn{1}{c}{Prec.} & \multicolumn{1}{c}{Rec.} & \multicolumn{1}{c}{\textit{F}} & (min) \\
        \midrule
        \multicolumn{3}{@{}l}{\textit{English (\%)}} \\
        JHU-PLP & 7.9 & \textbf{33.9} & 33.9 & 3.1 & 5.7 & 3.9 & 0.3 & 0.5 & 5.0 & 0.7 & 1.2 & - \\
        ES-KMeans & \textbf{100} & 72.6 & \textbf{51.0} & \textbf{54.4} & \textbf{52.7} & \textbf{13.0} & \textbf{14.1} & \textbf{13.5} & \textbf{8.3} & \textbf{16.7} & \textbf{11.1} & 788 \\
        \midrule
        \multicolumn{3}{@{}l}{\textit{French (\%)}} \\
        JHU-PLP & 1.6 & \textbf{25.4} & 30.9 & 0.6 & 1.1 & \textbf{5.2} & 0.1 & 0.1 & \textbf{6.9} & 0.2 & 0.3 & - \\
        ES-KMeans & \textbf{97.2} & 67.3 & \textbf{37.8} & \textbf{41.6} & \textbf{39.6} & 3.5 & \textbf{3.9} & \textbf{3.7} & {3.1} & \textbf{6.3} & \textbf{4.2} & 615\\
        \midrule
        \multicolumn{3}{@{}l}{\textit{Mandarin (\%)}} \\
        JHU-PLP & 2.9 & \textbf{30.7} & \textbf{37.5} & 0.9 & 1.8 & \textbf{4.0} & 0.1 & 0.1 & \textbf{4.5} & 0.1 & 0.2 & - \\
        ES-KMeans & \textbf{100} & {88.1} & {36.5} & \textbf{47.1} & \textbf{41.1} & {2.5} & \textbf{3.4} & \textbf{2.9} & {2.5} & \textbf{4.1} & \textbf{3.1} & 4\\
        \midrule
        \multicolumn{3}{@{}l}{\textit{Surprise Lang1 (\%)}} \\
        JHU-PLP & 3.0 & \textbf{30.5} & 28.2 & 1.2 & 2.3 & 4.0 & 0.1 & 0.2 & 5.5 & 0.3 & 0.6 & - \\
        ES-KMeans & \textbf{100} & {66.4} & \textbf{42.6} & \textbf{56.5} & \textbf{48.6} & \textbf{10.3} & \textbf{14.3} & \textbf{12.0} & \textbf{5.7} & \textbf{11.2} & \textbf{7.5} & 473 \\
        \midrule
        \multicolumn{3}{@{}l}{\textit{Surprise Lang2 (\%)}} \\
        JHU-PLP & 5.9 & \textbf{30.8} & 25.3 & 1.0 & 2.0 & 1.6 & 0.0 & 0.1 & 2.3 & 0.1 & 0.2 & - \\
        ES-KMeans & \textbf{100} & {72.2} & \textbf{42.4} & \textbf{44.3} & \textbf{43.3} & \textbf{4.9} & \textbf{5.2} & \textbf{5.0} & \textbf{4.6} & \textbf{10.0} & \textbf{6.3} & 5\\
        \bottomrule
    \end{tabularx}
    \endgroup
\end{table*}

When reading the results in Table~\ref{tbl:results}(b) from bottom to top,
note that
the results of BES-GMM do not converge to ES-KMeans, even though $\sigma^2$ is tending towards $0$.
This is because,
based on~\cite{murphy_bayesgauss07}, we set the variance of the prior on the component means of BES-GMM as $\sigma_0^2 = \sigma^2/\kappa_0$ ($\S$\ref{sec:experimental_setup}),
so
the prior variance on the component means is tied to the fixed data variance.
Under these conditions,
the asymptotic equivalence of BES-GMM and ES-KMeans 
no longer holds.
Murphy~\cite{murphy_bayesgauss07} explains that this coupling is a sensible way to incorporate prior knowledge of the typical spread of data,
and here we indeed show how this helps our Bayesian model; this principled way of including priors is not possible in ES-KMeans or SylSeg.
When setting $\sigma_0^2 = 1$ (rather than tying it), the results of BES-GMM match ES-KMeans when $\sigma^2 =0.00001$.

\subsection{Results: Zero Resource Speech Challenge 2017}

Next, we use ES-KMeans as our entry to ZRSC'17.
For the challenge, data for three languages were released in advance with evaluation code, allowing participants to develop their systems. 
After registering their optimal hyperparameters, participants could apply their systems to two surprise languages and submit their system output to the organizers for evaluation.
Thus, the challenge measures how well systems and hyperparameters generalize to completely unseen languages.
The data consists of
an English corpus of about 45\ hours of speech from 69 speakers, a French corpus of 24 hours from 28 speakers, a Mandarin corpus of 2.5 hours from 12 speakers, the surprise Lang1 corpus of 25 hours from 30 speakers, and the surprise Lang2 corpus of about 10 hours of speech from 24 speakers.\footnote{\url{http://sapience.dec.ens.fr/bootphon/2017/}}

{
To evaluate whether ES-KMeans scales, and in the spirit of the challenge which explicitly want to measure robustness to multiple speakers, we use a speaker-independent setup here.\footnote{BES-GMM can also be applied to multiple speakers, as in~\cite{kamper+etal_arxiv16}. However, on the 5-hour English data set of $\S$\ref{sec:results_zs2015}, the speaker-independent BES-GMM takes around 40 hours to train.
Since we wanted to compare different models and hyperparameters in a reasonable time, we used speaker-dependent modelling in $\S$\ref{sec:results_zs2015}.
Furthermore, SylSeg is also speaker-dependent~\cite{rasanen+etal_interspeech15}.}
To deal with the increased computational load, we use more constrained hyperparameters than in the previous section: $K$ is set to $10\textrm{\%}$ of the number of first-pass segmented syllables, and word candidates are limited to at most 4 syllables.
We also parallelize ES-KMeans over 15~CPUs.
The only hyperparameter we explicitly tried to optimize was the probability with which word boundaries are initialized; the optimal values varied across languages and metrics, so we just randomly initialize boundaries with a 0.5 probability.
}

{
Table~\ref{tbl:results_zs2017} compares the performance of ES-KMeans to the challenge baseline---an unsupervised term discovery system (JHU-PLP)---on the 5 languages.
Term discovery systems like JHU-PLP~\cite{jansen+vandurme_asru11} aim to find high-precision clusters of isolated segments, but do not cover all the data.
Compared to ES-KMeans, it therefore performs much better on NED, which only evaluates the discovered patterns. 
In contrast to JHU-PLP, ES-KMeans performs full-coverage segmentation.
It therefore achieves better recall, \textit{F}-score and coverage performance on all 5 languages.
For word boundary, token and type precision, JHU-PLP performs better on some of the Mandarin and French metrics, but ES-KMeans scores better on the~other~languages.
}

{
Apart from word boundary detection, the scores of all metrics in Table~\ref{tbl:results_zs2017} are very low for both systems, showing that zero-resource speech processing is still far from mature.
Nevertheless, hyperparameters seem to generalize across languages, with ES-KMeans performing better on Lang1 and Lang2 than on either French or Mandarin.
Moreover, ES-KMeans can be trained in a reasonable time on realistically sized corpora, taking 13 hours to train on the 45-hour English corpus.
}

\section{Conclusion}

We introduced the embedded segmental \textit{K}-means model (ES-KMeans), a method that falls in between the fully Bayesian embedded segmental GMM (BES-GMM) and the cognitively motivated heuristic SylSeg method.
Its word segmentation performance is on par with BES-GMM and superior to SylSeg, but cluster purity is worse than both other methods.
In terms of efficiency, it is 5 times faster than BES-GMM, but half as fast as SylSeg.
Despite using hard clustering and segmentation, ES-KMeans still has a clear objective function and it is guaranteed to converge (to a local optimum), in contrast to SylSeg.
It also has far fewer hyperparameters than BES-GMM, although we show that this is what gives the latter the upper hand.
{Due to its improved efficiency, we are also able to apply ES-KMeans to larger corpora, on which it performs competitively.}
{In future work we will focus on improving the acoustic word embedding function, which ES-KMeans could directly incorporate since it is agnostic to the embedding method.}

\vspace*{4pt}
\noindent\textbf{Acknowledgements:} We would like to thank Okko R{\"a}s{\"a}nen and Shreyas Seshadri for helpful feedback on the  SylSeg~model.\newpage

\balance
\bibliography{asru2017}

\end{document}